\documentclass[10pt,twocolumn,letterpaper]{article}

\usepackage{wacv}
\usepackage{times}
\usepackage{epsfig}
\usepackage{graphicx}
\usepackage{amsmath}
\usepackage{amssymb}

\usepackage{booktabs}
\usepackage{tabularx}
\newcolumntype{Y}{>{\centering\arraybackslash}X}

\newcommand{\veryshortarrow}[1][3pt]{\mathrel{%
		\hbox{\rule[\dimexpr\fontdimen22\textfont2-.2pt\relax]{#1}{.4pt}}%
		\mkern-4mu\hbox{\usefont{U}{lasy}{m}{n}\symbol{41}}}}
\newcommand{\adapt}[2]{\textit{#1} $\veryshortarrow$ \textit{#2}}

\newcommand{\tableheaderadapt}[3]{\midrule\multicolumn{#1}{c}{\adapt{#2}{#3}} \\ \midrule}
\usepackage[nomessages]{fp}
\newcommand\Ts{\rule{0pt}{2.6ex}}         
\newcommand\Bs{\rule[-0.9ex]{0pt}{0pt}}   
\newcommand\cityscapescityscapes{63.6}
\newcommand\cityscapesmapillary{43.2}

\newcommand\cityscapesapollo{25.8}

\newcommand\gtacityscapes{34.8}
\newcommand\gtagta{62.1}
\newcommand\gtamapillary{37.1}
\newcommand\gtaapollo{25.3}

\newcommand\gtappgta{61.2}
\newcommand\gtappcityscapes{22.1}
\FPeval{\differencepp}{round(\gtacityscapes - \gtappcityscapes, 1)}

\newcommand\gtappcustomgta{61.9}
\newcommand\gtappcustomcityscapes{35.0}

\newcommand\gtanobncityscapes{26.5}

\newcommand\uadacityscapes{38.2}
\newcommand\uadagta{62.7}
\newcommand\uadamapillary{38.5}
\newcommand\uadaapollo{27.4}
\FPeval{\gapredcustom}{round(\uadacityscapes - \gtacityscapes, 1)}
\FPeval{\gapredcustommapillary}{round(\uadamapillary - \gtamapillary, 1)}
\FPeval{\gapredcustomapollo}{round(\uadaapollo - \gtaapollo, 1)}

\newcommand\uadainstancecityscapes{31.4}
\newcommand\gtainstancecityscapes{30.3}
\FPeval{\gapredinstance}{round(\uadainstancecityscapes - \gtainstancecityscapes, 1)}

\newcommand\uadasplitcityscapes{35.4}
\newcommand\gtasplitcityscapes{\gtacityscapes}
\FPeval{\gapredsplit}{round(\uadasplitcityscapes - \gtasplitcityscapes, 1)}

\newcommand\uadanobncityscapes{28.4}
\FPeval{\gaprednobn}{round(\uadanobncityscapes - \gtanobncityscapes, 1)}

\newcommand\uadamultibatchcityscapes{29.6}
\FPeval{\gapredmultibatch}{round(\uadamultibatchcityscapes - \gtacityscapes, 1)}

\newcommand\uadatwomapillary{45.0}
\newcommand\uadatwocityscapes{64.0}
\newcommand\uadatwoapollo{27.1}
\FPeval{\gapreduadatwo}{round(\uadatwomapillary - \cityscapesmapillary, 1)}
\FPeval{\gapreduadatwoapollo}{round(\uadatwoapollo - \cityscapesapollo, 1)}

\wacvfinalcopy 

\def\wacvPaperID{37} 

\ifwacvfinal\fi
\begin{document}

\title{A Domain Agnostic Normalization Layer \\ for Unsupervised Adversarial Domain Adaptation}

\author{R. Romijnders \hspace{2cm} P. Meletis \hspace{2cm} G. Dubbelman \\
Eindhoven, University of Technology\\
{\tt\small romijndersrob@gmail.com}
}

\maketitle
\ifwacvfinal\thispagestyle{empty}\fi

\begin{abstract}
   We propose a normalization layer for unsupervised domain adaption in semantic scene segmentation. Normalization layers are known to improve convergence and generalization and are part of many state-of-the-art fully-convolutional neural networks. We show that conventional normalization layers worsen the performance of current Unsupervised Adversarial Domain Adaption (UADA), which is a method to improve network performance on unlabeled datasets and the focus of our research. Therefore, we propose a novel Domain Agnostic Normalization layer and thereby unlock the benefits of normalization layers for unsupervised adversarial domain adaptation. In our evaluation, we adapt from the synthetic GTA5 data set to the real Cityscapes data set, a common benchmark experiment, and surpass the state-of-the-art. As our normalization layer is domain agnostic at test time, we furthermore demonstrate that UADA using Domain Agnostic Normalization improves performance on unseen domains, specifically on Apolloscape and Mapillary.\footnote{Accepted to IEEE WACV 2019}
\end{abstract}

\section{Introduction}

%




Semantic segmentation constitutes a crucial task in computer vision of assigning a class to every pixel in an image. Applications range from autonomous driving and robotic navigation to segmenting natural scenes. Convolutional neural networks (CNNs) have shown good performance, \eg \cite{Badrinarayanan2015,Pathak2014}. However, when we train these models on a \textit{source domain}, and evaluate on another, \textit{target domain}, the performance degrades. Ideally, we would adapt our model to the target domain, without requiring to label the images (unsupervised). In this work, we aim for unsupervised domain adaptation to improve generalization capability for semantic scene segmentation. 

\par 
Unsupervised domain adaptation addresses three problems: 1) In new domains, images differ in appearance, lighting, contrast or colorization. For example, when the time of day, season or camera changes. This shift is studied in \cite{Khosla2012,Torralba2011,tommasi2015deeper}. We want good performance for all domains; 2) Labeling the data is a cumbersome task. For example, labeling the Cityscapes data set took up to 90 minutes per image \cite{Cordts2016}. Unsupervised domain adaptation alleviates this labeling burden as no labels are required for the target domain; 3) Models trained on simulated data fail to perform well on real data. Domain adaptation poses a potential solution to close this gap.

\begin{figure}  
	\centering
	\includegraphics[width=\linewidth]{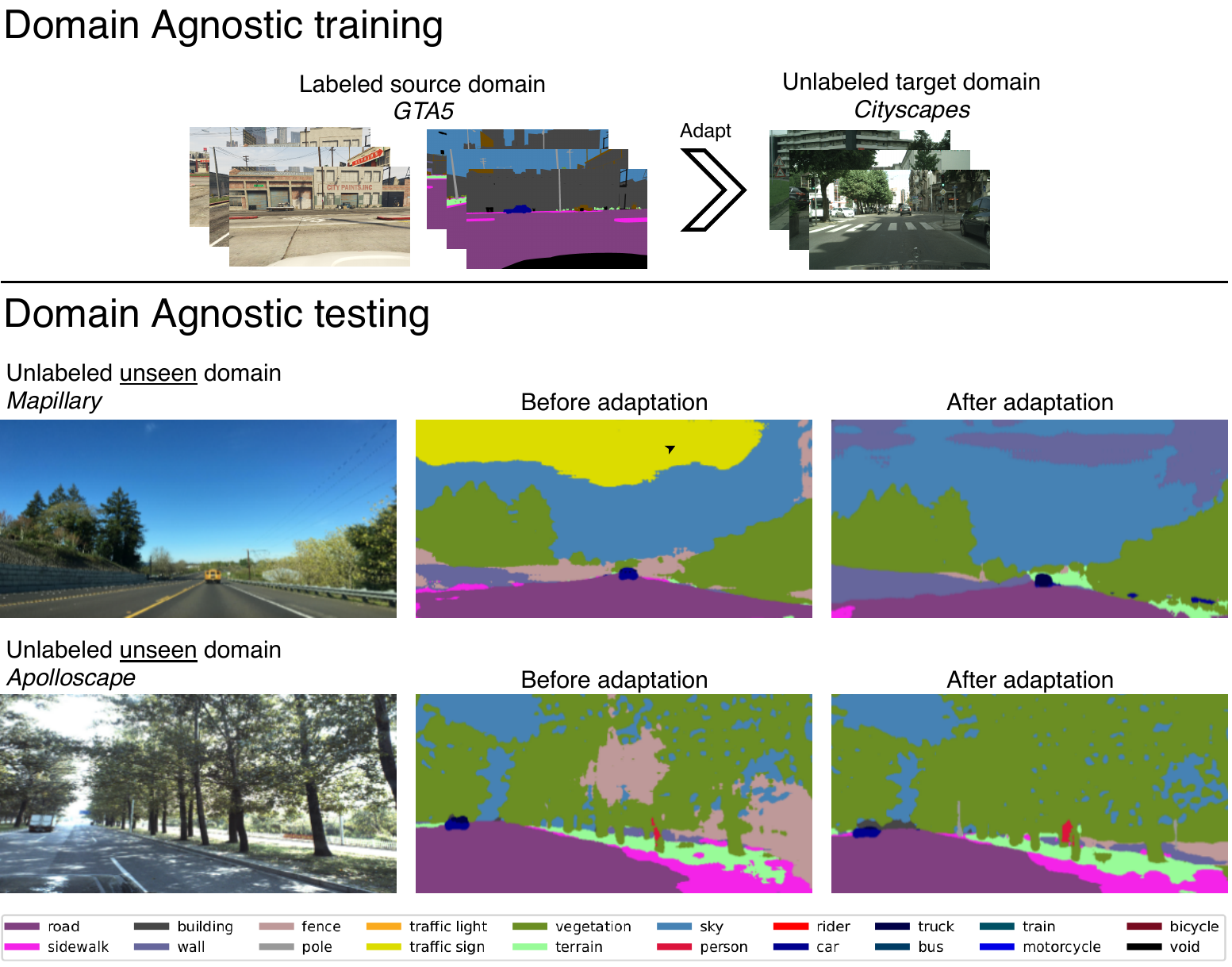}
	\caption{Diagram for unsupervised domain adaptation. During training, we only have labels for the source domain. During testing, we want a model that we can apply to any unseen and unlabeled domain, without any domain-specific fine tuning.}
	\label{opening_figure}
\end{figure}

\par 
A recent and promising work in unsupervised domain adaptation for semantic segmentation is unsupervised adversarial domain adaptation (UADA) \cite{Ganin2017}, which will be the focus of our work. Current approaches in UADA have two challenges. First, evaluations of the model focus on the source and target domain, but omit any unseen domains. For example, when we adapt a segmentation model from Germany to China, we also want good performance in other Asian countries. To summarize, we aim to improve the generalization capability of the model to any (unseen) domain, not limiting to only the source and target domain. 

\par 
As a second challenge, many recent approaches in UADA use CNNs without normalization layers. This is suboptimal as normalization layers speed up convergence and reduce sensitivity to initialization of the parameters and hyperparameters \cite{ioffe2015batch,towardstheoryBN,howdoesBN,understandingBN}. For example, the winning entries of the recent ImageNet competition and Robust Vision competition at CVPR 2018 use normalization layers \cite{rob1,bae2017rank,xie2017aggregated,hu2017squeeze}. However, none of previous work in UADA use normalization layers \cite{UnsupGAN,FCNWild,NoMore}. This absence probably follows from observations that normalization layers reduce performance during adversarial training \cite{salimans2016improved, xiang2017effects}. 


\par  
Batch normalization \cite{ioffe2015batch}, the most commonly used normalization layer, makes the output for one image dependent on another image in a batch. This dependency creates havoc when the batch contains images from both labeled and unlabeled domains. Therefore, we propose a new normalization layer for UADA. In Section \ref{results:normalization_layer}, we show that our normalization layer does not degrade performance in UADA, as conventional normalization layers do. Next, in Section \ref{results:UADA}, we show that we surpass state-of-the-art, \cite{FCNWild}, adaptation using our normalization layer. Moreover, we show that this adaptation also improves performance on two unseen domains.

\par 
We evaluate our approach on large-scale scientific benchmarks for semantic scene segmentation. We experiment on data from urban scenes as these data pose many challenges due to object clutter and have a wide diversity of classes and appearance. We consider two adaptation tasks: from synthetic to real data, GTA5 to Cityscapes; from real to real data, Cityscapes to Mapillary \cite{Cordts2016,Richter2016,neuhold2017mapillary}. To the best of our knowledge, we are the first to evaluate performance on unseen domains, Mapillary and Apolloscape \cite{neuhold2017mapillary,apolloscape}.

\par 
As we unlock the benefits of normalization layers for UADA, we make the following contributions:

\begin{itemize}
	\item We demonstrate a degraded performance when using conventional batch normalization in UADA and provide insight in this degradation using an experiment. (Section \ref{results:normalization_layer})
	\item We propose a Domain Agnostic Normalization layer for UADA (Section \ref{method:normalization_layer}), surpass state of the art, \cite{FCNWild}, and show a performance improvement on two unseen domains (Section \ref{results:UADA}).
\end{itemize}
Our code to reproduce the results is available at \cite{code_repo}.

\section{Related work}

\subsection{Semantic segmentation}

\par 
Semantic segmentation has been widely studied in computer vision. Many current semantic segmentation models use CNNs, following the progress in large-scale image classification \cite{farabet2013learning, hariharan2014simultaneous}. Neural networks learn hierarchical representations using layers of neurons \cite{Bengio2013}. 


\par 
Recently, fully convolutional networks (FCN) generalize CNNs for arbitrary input sizes \cite{Long_fcn}. This generalization enables the use of classification networks such as VGG \cite{simonyan2014very} or ResNet \cite{he2016deep} for semantic segmentation. In this generalization, each layer comprises a grid of individual representations, one representation for each receptive field. As semantic segmentation deals with large images, \cite{Yu2016} introduced dilated convolutions to enlarge the receptive field and the authors of \cite{Badrinarayanan2015,Noh2015} proposed to learn the upsampling for large label maps.


\subsection{Adversarial adaptation}

\par 
Adversarial adaptation draws inspiration from Generative Adversarial Networks (GAN) \cite{Goodfellow2014}. In training a GAN, a generator outputs data samples while trying to confuse a discriminator that classifies between generated and real data samples. 

\par 
Analogous to using a discriminator to align generated and real data samples, adversarial training uses an adversary to align representations from multiple domains in a neural network. In other words, a domain classifier serves as an adversary and learns to discriminate source from target representations. Another model learns to confuse the domain classifier, which encourages alignment between the source and target representations. As the representations from multiple domains become more aligned, the neural network will generalize better to new domains \cite{Ben-David2010}.

\subsection{Domain adaptation}



\par 
Ganin et al. \cite{Ganin2017} propose adversarial adaptation to align the representations. This work has been extended in \cite{Tzeng_2015_ICCV} and \cite{ADDA} for image classification. Another adversarial approach to domain adaptation is to transfer the style of an image from source to target domain \cite{NIPS2016_6544, bousmalis2017unsupervised, zhu2017unpaired}.

UADA has been applied to semantic segmentation. For example, \cite{FCNWild} builds on the work of \cite{Tzeng_2015_ICCV} and applies a domain classifier on the representations learned by an FCN. UADA aims for alignment at the representation level. Later works have focused on combining alignment at the representation level with alignment at the logit level, \cite{NoMore}, or alignment at the output level, \cite{UnsupGAN}. We focus on UADA for semantic segmentation models that use normalization layers. Normalization layers pose specific challenges for UADA that we will address in this work.

\subsection{Normalization}
\par 
Using normalization layers in a neural network speeds up training and reduces sensitivity to initialization of the parameters and hyperparameters \cite{ioffe2015batch,towardstheoryBN,howdoesBN,understandingBN}. 

Perhaps the earliest attempt was local response normalization by Krizhevsky et al. \cite{krizhevsky2012imagenet}. Ioffe and Szegedy introduced the batch normalization layer that normalizes all representations in a batch \cite{ioffe2015batch}. During training, the internal activations in the many layers may shift. Batch normalization normalizes the inputs to zero mean and unit variance.

\par Batch normalization has the disadvantage that predictions depend on all samples in a batch. To alleviate this dependency, other normalization layers, such as instance normalization \cite{instancenorm} and layer normalization \cite{ba2016layer} are proposed. For a representation grid of size $N (\text{Sample size}) \times H (\text{Height}) \times W (\text{Width}) \times C (\text{Channels})$ in a CNN, these layers normalize over the following axes:
\begin{itemize}
	\item \textit{Batch normalization} normalizes over $N, H, W$
	\item \textit{Instance normalization} normalizes over $H, W$
	\item \textit{Layer normalization} normalizes over $H, W, C$
\end{itemize}
\par 

In this work, we compare our proposed normalization layer with two forms of batch normalization and with instance normalization. We do not consider layer normalization, as batch normalization is known to outperform layer normalization in CNNs \cite{ba2016layer}. In Section \ref{method:normalization_layer}, we propose our Domain Agnostic Normalization, compare to other layers in Section \ref{results:normalization_layer}, and show the benefits for domain agnostic testing in Section \ref{results:UADA}.

\section{Method: Domain Agnostic Normalization for unsupervised domain adaptation} \label{method:section}

\par 
In this section, we present the details of our proposed Domain Agnostic Normalization layer. First, in Section \ref{method:domain_adaptation}, we introduce notation and set up UADA. Second, in Section \ref{method:normalization_layer}, we address the problems of conventional normalization in UADA and propose Domain Agnostic Normalization.

\subsection{Domain adaptation} \label{method:domain_adaptation}

\begin{figure}[t]
	\centering
	\includegraphics[width=\linewidth]{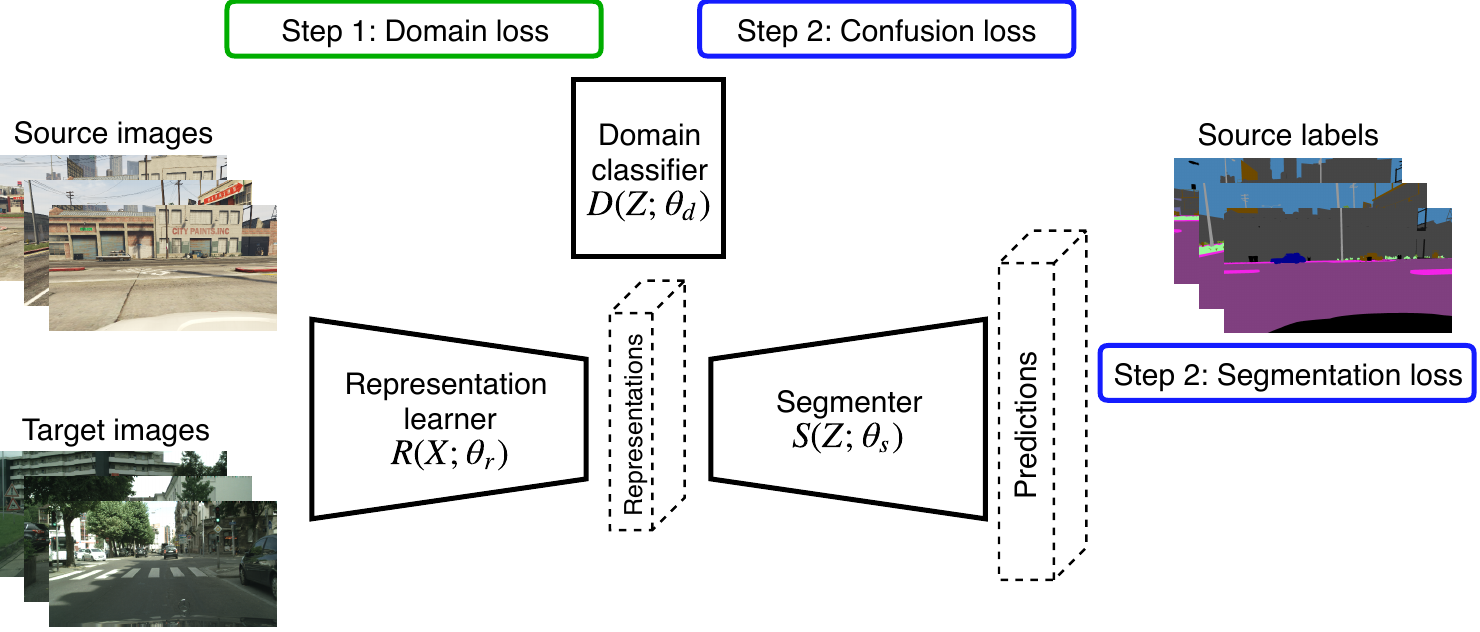}
	\caption{Diagram depicting the two alternating steps during UADA. In the first step, we train the domain classifier using the domain loss. In the second step, we train the segmentation model using the segmentation and confusion loss. (see Equation \ref{eq_alternating})}
	\label{UADA_system}
\end{figure}

\par 
We aim to learn a segmentation model that takes an image, $X \in \mathbb{R}^{H \times W \times C}$, and segments the image to a label map, ${ \hat{Y} \in \mathbb{R}^{H \times W \times K} }$. Here, $C$ is the number of input channels, $K$ is the number of output classes and $H$ and $W$ are height and width, respectively. We train using labeled source data, ${(X_s, Y_s)}$, and  unlabeled target data, ${X_t}$. 

\par 
Figure \ref{UADA_system} displays a diagram of the UADA model. The model consists of three major blocks. First, the representation learner parametrizes the representations using a CNN, $Z = R(X;\theta_{r})$. This representation learner can contain an arbitrary number of normalization layers. We learn the parameters of the representation learner using both the source (labeled) and target (unlabeled) domain. Second, the segmenter maps the representations to a class prediction for each pixel, $\hat{Y} = S(Z;\theta_{s})$. We learn the parameters of the segmenter using the source domain. We assume that all domains have the same classes. So during testing, we re-use the segmenter on any domain (Domain Agnostic testing, Figure \ref{opening_figure}). Finally, the domain classifier assigns a probability, $p$, that a representation comes from an image in the target domain, $p = D(z; \theta_{d})$. We learn the parameters of the domain classifier using the images from both the source and target domain.

\par Training UADA requires two alternating steps as confusing the representations opposes the learning of a domain classifier. In the first of alternating steps, we minimize the domain loss, $\mathcal{L}_{dom}(X_s, X_t; \theta_r, \theta_d)$, to learn the parameters of the domain classifier. As we have two domains, the domain loss is a binary cross entropy loss. In the second of alternating steps, we minimize the segmentation loss, $\mathcal{L}_{segm}(X_s, Y_s; \theta_r, \theta_s)$, and the confusion loss, $\mathcal{L}_{conf}(X_t; \theta_r, \theta_d)$. The segmentation loss is a multi-class cross entropy for the $K$ output classes. The confusion loss encourages the segmentation model to confuse a target representation for a source representation. In other words, the confusion loss assumes a low value when the domain classifier predicts a low value of $p$ for any target representation. Therefore, the confusion loss follows:
\begin{equation}
\mathcal{L}_{conf}(X_t; \theta_r, \theta_d) = -\sum_{z \in R(X_t; \theta_r)} \log (1-D(z; \theta_d))
\label{conf_loss_target}
\end{equation}

\par 
In total, we train unsupervised adversarial domain adaptation using the following alternating steps:
\begin{equation}
\begin{aligned}
& \min_{\theta_{d}}  \ \ \mathcal{L}_{dom} (X_s, X_t; \theta_r, \theta_d)                                                                  &   \\ 
& \min_{\theta_{r}, \theta_s} \ \  \mathcal{L}_{segm}(X_s, Y_s; \theta_r, \theta_s) + \lambda\mathcal{L}_{conf}(X_t; \theta_r, \theta_d) &   
\end{aligned}
\label{eq_alternating}
\end{equation}
$\lambda$ trades off the segmentation loss and the confusion loss. Figure \ref{UADA_system} displays a diagram for this alternating scheme.

\subsection{Domain Agnostic Normalization layer} \label{method:normalization_layer}


\begin{figure}[t]
	\centering
	\includegraphics[width=\linewidth]{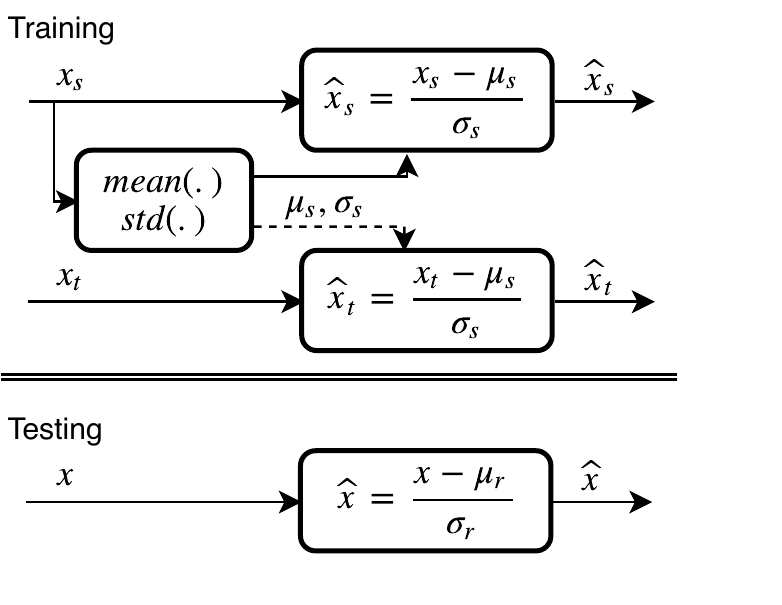}
	\caption{Diagram for our Domain Agnostic Normalization layer. We transform the source representations, $x_s$, using the source statistics, $\mu_s$ and $\sigma_s$. We transform the target domain and all unseen domains using the source statistics as fixed parameters. Hence, the dotted line indicates that no gradients flow during back propagation.}
	\label{fig:batch_norm_split}
\end{figure}
\par 
In this section, we propose our Domain Agnostic Normalization (DAN) layer for UADA. First, we outline a major problem with batch normalization, the most commonly used normalization layer \cite{ioffe2015batch}. Second, we address these problems and propose DAN.

\par 
Batch normalization introduces dependencies between all representations in a batch. For each batch, batch normalization calculates the statistics on all representations and applies these statistics to any representation. This dependency makes the supervision on any image depend on the representations of any other image in the batch. When doing multi-domain training, batches contain images of multiple domains. Now this dependency makes the supervision on one domain depend on representations in another domain. For multi-domain training using only labeled domains, this dependency poses no problem. For a representation in one image, the supervision from its own prediction would be stronger than the implicit supervision of other images via the statistics. For multi-domain training also using an unlabeled domain, as in UADA, however, this dependency creates havoc on the representations of the unlabeled domain. These representations get only supervision from other images via the statistics and have no supervision from a prediction to outweigh this influence. Such a learning procedure would counteract segmentation performance on the target domain. Our experiments in Section \ref{experiments} will confirm this detrimental effect.

\par 
The incompatibility of conventional batch normalization and UADA motivates our requirements for a new normalization layer:
\begin{enumerate}
	\item \label{desi_no_dep} A normalization layer that introduces no dependency between source and target representations. Learning representations can only depend on information from one domain. 
	\item \label{desi_same_trans} A normalization layer that applies the same transformation to any domain. Minimizing the confusion loss in Equation \ref{conf_loss_target} should concern the differences in learned representations and not the differences in normalization transformations. Moreover, as we use the same transformation on any domain, we do not need to fine-tune our model when we test on unseen domains (Domain Agnostic testing, Figure \ref{opening_figure}).
\end{enumerate}


\par 
To satisfy these requirements, we propose our Domain Agnostic Normalization layer. Figure \ref{fig:batch_norm_split} shows a diagram of DAN. During training, DAN makes the following steps:
\begin{enumerate}
	\item For the \textit{source domain}, our normalization layer operates like batch normalization. We calculate the statistics on all representations of the source domain per batch and transform the representations using these statistics.
	\item For the \textit{target domain}, our normalization layer transforms the representations using the statistics from the source as fixed parameters. This fixation of parameters causes no gradients to flow during backpropagation, so our normalization layer introduces no dependencies across domains.
\end{enumerate}
\par 
During testing, we use the same transformation on all domains, the source domain, the target domain, and any unseen domains. This transformation uses the statistics for the source domain that we aggregate during training. Other works have proposed to re-estimate the statistics for each new domain \cite{li2016revisiting}. Our DAN remains agnostic to the domain at test time and does not require the additional effort of re-estimating the statistics.

\section{Experiments} \label{experiments}
In this section, we outline three sets of experiments that demonstrate the benefit of DAN in UADA. 

\par 
Our first set of experiments compares the alternatives for DAN in UADA. We run experiments using four normalization layers:
\begin{itemize}
	\item \textit{Conventional batch normalization} introduces a dependency between representations from source and target domain. When the target domain has no labels, we expect that batch normalization decreases the performance on the target domain. 
	\item \textit{Split batch normalization \cite{tipsGAN}} applies one conventional batch normalization layer per domain. This split removes the dependencies between domains. However, split batch normalization does not apply the same transformation to each domain. 
	\item \textit{Instance normalization \cite{instancenorm}} normalizes the representations per image and introduces no dependencies at all. Instance normalization, however, calculates the statistics using $H \times W$ representations per layer instead of the $N \times H \times W$ representations that batch normalization uses. This smaller sample introduces more stochasticity.
	\item \textit{Domain Agnostic Normalization} introduces no dependencies between representations across domains and applies the same transformation in each domain.
\end{itemize}

Our second set of experiments demonstrates the benefits of DAN on two adaptation tasks:
\begin{itemize}
	\item Adaptation from synthetic to real data, from GTA5 to Cityscapes. We run this experiment to confirm that we reproduce state of the art in UADA, \cite{FCNWild,NoMore,UnsupGAN}. We compare to these works in Section \ref{results:normalization_layer}.
	\item Adaptation from real to real data, from Cityscapes to Mapillary. So far as we know, we are the first work to report on this adaptation task. We value this adaptation task as Cityscapes is shot in one country using one camera, whereas Mapillary is shot in multiple countries using multiple cameras (Section \ref{experiments:datasets} provides the details per data set).
\end{itemize}
\par 
Our third set of experiments evaluates the final models of both adaptation tasks on unseen domains. The main benefit of DAN occurs in domain agnostic testing. DAN acts as a fixed transformation during testing. One might wonder if this fixation generalizes to unseen domains. Therefore, we evaluate the segmentation model that we adapt from GTA5 to Cityscapes on two unseen domains, Mapillary and Apolloscape. We evaluate the segmentation model that we adapt from Cityscapes to Mapillary on the unseen domain, Apolloscape. To the best of our knowledge, we are the first to evaluate these adaptation tasks on unseen domains.

\subsection{Data sets}\label{experiments:datasets}

\par 
\textbf{Cityscapes}\cite{Cordts2016} contains images from 50 cities in and around Germany. Images show scenery over several months, all in good weather conditions. The dense labels consist of nineteen categories of the urban scene, \eg road, sidewalk, car, bus and traffic sign. All images are shot using the same car and camera. Sample sizes: training, 2975; validation, 500. 
\par 
\textbf{GTA5}\cite{Richter2016} contains images rendered from the Grand Theft Auto computer game. Images are taken in the fictitious city Los Santos. The dense labels follow the class definitions of the Cityscapes data set. Sample sizes: training, 9000; validation, 3000.
\par 
\textbf{Mapillary}\cite{neuhold2017mapillary} contains images from all around the world. Images are taken in varying weather conditions, seasons, and time of day. Images are shot using different devices such as mobile phones, action cameras and professional equipment. The dense labels cover 66 classes, but we use only the 19 classes that overlap with Cityscapes and GTA5. Sample sizes: training, 18000; validation, 2000.
\par 
\textbf{Apolloscape}\cite{apolloscape} contains images from Beijing, China. Images are taken in bright weather conditions and are shot using the same car and camera. We use only the 19 classes that overlap with Cityscapes and GTA5. The labeling process was partially automated, which introduces artefacts. Sample size: we use 8327 images from the validation set.

\subsection{Performance metric}\label{method:perf}

\par 
We follow the evaluation in \cite{FCNWild,UnsupGAN,Zhang2017} and report results on $19$ classes. We consider mean intersection over union (mIOU) as our figure of merit. Intersection over union (IOU) represents the number of pixels predicted correctly per class divided by the total number of predicted and true pixels for that class. mIOU averages the IOU over all classes. Perfect segmentation would achieve 100 \% mIOU.
\par 
We assess the confusion of source and target representations using a retrieval curve, as in \cite{UnsupGAN}. Per target representation, we consider a number of nearest representations (horizontal axis) and average the number of source representations retrieved (vertical axis). Perfect alignment occurs when half of the nearest representations are source representations and half of the nearest representations are target representations. In other words, for any $m$ target representations, perfect alignment occurs when $\frac{m}{2}$ representations come from the source domain and $\frac{m}{2}$ representations come from the target domain. The more two sets of representations are aligned, the closer their retrieval curve lies to the perfect alignment curve. In all retrieval curves in this work, we sample five thousand source and five thousand target representations from images in the respective validation sets.

\subsection{Implementation details} \label{implementation_details}
\par 
In all models, the segmentation model trains for 17 epochs. Adversarial adaptation learns from two data sets at the same time. We define an epoch when, in expectation, we sampled the smallest data set once. Due to computational limits, we resize all images and labels to size $384 \times 768$. For the purposes of this work, we consider a new data set to be a new domain.
\par 
For training, we use stochastic gradient descent with momentum. Learning rate and momentum are set at $0.01$ and $0.9$, respectively. We half the learning rate after 9 and 16 epochs. The batches contain two images per domain. The network uses pre-trained parameters from ResNet-50 \cite{he2016deep}, trained on ImageNet \cite{deng2009imagenet}. To stabilize the training of the domain classifier, we use instance noise \cite{arjovsky2017towards}. All experiments run in Tensorflow 1.6 \cite{Abadi2015}.

\section{Results} \label{results}
\subsection{Comparing normalization layers} \label{results:normalization_layer}
\begin{table}
	\caption{\textbf{Analyzing batch normalization on multi-domain training.} The units are \% mIOU. The \textit{batch constituents} column indicates the domains of images in the batch, G refers to GTA5 and C refers to Cityscapes. In this experiment, we train models using only the GTA5 labels.}
	\label{table__1plusplus}
	\centering
	\begin{tabular}{l||l|c|c}
		Normalization  & Batch        & \multicolumn{2}{c}{Tested on} \\ 
		layer      & constituents & GTA5             & Cityscapes                     \Bs \\
		\hline
		Batch norm & (G, G)       & \gtagta          & \gtacityscapes                 \Ts \\
		Batch norm & (G, G, C, C) & \gtappgta        & \gtappcityscapes                \\
		DAN        & (G, G, C, C) & \gtappcustomgta  & \gtappcustomcityscapes \\
	\end{tabular}
\end{table}

\begin{table}
	\caption{\textbf{Comparing normalization layers on adaptation \adapt{GTA5}{Cityscapes}.} Numbers represent performance on the target domain. Gap reduction indicates the difference between source-only and UADA. Norm. abbreviation normalization. All units are \% mIOU.}
	\label{table__2normlayers}
	\centering
	\setlength\tabcolsep{4pt}
	\begin{tabularx}{\linewidth}{r|Y|Y|Y}
		Normalization layer        & \multicolumn{1}{p{1.2cm}|}{\centering Source\\ only} & UADA & \multicolumn{1}{p{1.3cm}}{\centering Gap \\ reduction} \Bs \\ 
		\hline
		No norm. & \gtanobncityscapes					  & \uadanobncityscapes      & \gaprednobn           \Ts \\
		Batch norm. \cite{ioffe2015batch} & \gtacityscapes & \uadamultibatchcityscapes & \gapredmultibatch \\
		Instance Norm. \cite{instancenorm}& \gtainstancecityscapes                       & \uadainstancecityscapes  & \gapredinstance        \\
		Split batch norm.  \cite{tipsGAN}& \gtasplitcityscapes                          & \uadasplitcityscapes     & \gapredsplit           \\
		DAN [ours] & \textbf{\gtacityscapes}  & \textbf{\uadacityscapes} & \textbf{\gapredcustom} \\
	\end{tabularx} %
\end{table}

\begin{figure}[t]
	\centering
	\includegraphics[width=\linewidth, clip=true, trim = 0mm 0mm 0mm 23mm]{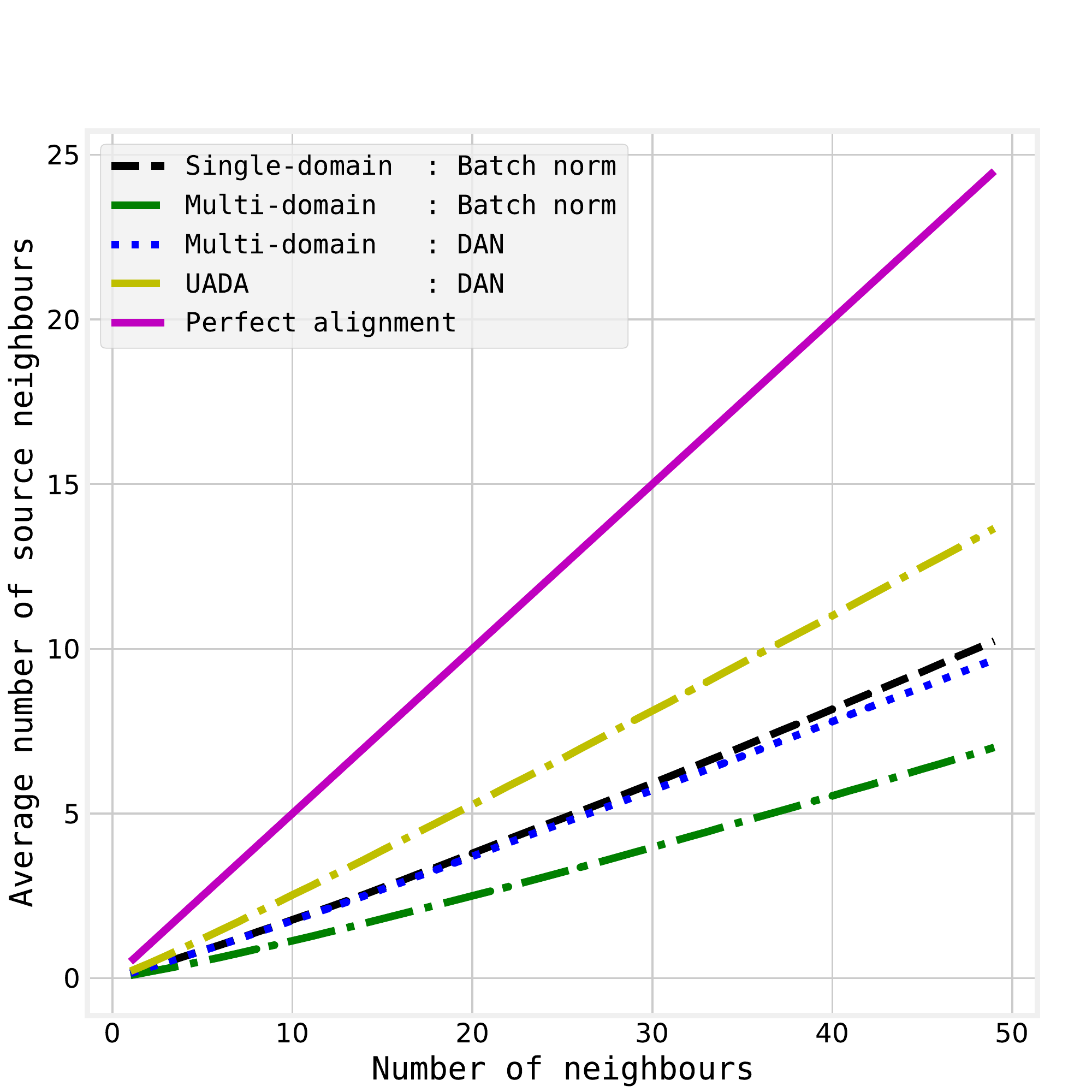}
	\caption{Retrieval curve to show the effect of conventional batch normalization on the representations. Per target representation, we consider a number of nearest representations (horizontal axis) and average the number of source representations retrieved (vertical axis). Perfect alignment occurs when half of the nearest representations are source representations and half of the nearest representations are target representations; the more aligned the representations are, the closer their curve lies to the perfect alignment curve. We explain retrieval curves in Section \ref{method:perf}}
	\label{retrieval_plot}
\end{figure}

\par 
First, we report experimental results on single domain training using conventional batch normalization and our normalization layer. In Section \ref{method:normalization_layer}, we reasoned that conventional batch normalization degrades the performance on the unlabeled target domain in unsupervised domain adaptation. This degradation is undesirable for UADA, where we aim for segmentation performance on the target domain. 

\par Table \ref{table__1plusplus} reports the results for this experiment. A source-only model on GTA5 using conventional batch normalization achieves $\gtagta$ \% mIOU on the GTA5 and $\gtacityscapes$ \% mIOU on Cityscapes. When we include the unlabeled images from Cityscapes in the batches, the performance changes to $\gtappgta$ \% mIOU on GTA5 and $\gtappcityscapes$ \% mIOU on Cityscapes. These results confirm that conventional batch normalization degrades the performance on the unlabeled target domain, Cityscapes, by $\differencepp$ points mIOU. When we change the conventional batch normalization layer to our DAN, the model achieves $\gtappcustomgta$ \% mIOU on GTA5 and $\gtappcustomcityscapes$ \% mIOU on Cityscapes. These numbers are comparable to single domain training and confirm that multi-domain training using our DAN does not degrade the performance on the unlabeled target domain. 

\par 
The retrieval curve in Figure \ref{retrieval_plot} shows another view on the effects of batch normalization in unsupervised domain adaptation. We compare the retrieval curves for using conventional batch normalization or DAN when we include images from an unlabeled domain in the batch. The curve for conventional batch normalization in multi-domain training (green) lies below the source-only curve (black) and the curve for multi-domain training using DAN (blue). This difference shows that conventional batch normalization learns representations for the unlabeled domain that are less aligned with the representations from the labeled domain. This decrease in alignment opposes unsupervised domain adaptation, where we aim for perfect alignment of the representations in both domains.

\par 
Next, we compare DAN with four alternatives: conventional batch normalization, no normalization, split batch normalization, \cite{tipsGAN}, and instance normalization, \cite{instancenorm}. Table \ref{table__2normlayers} shows the results for UADA using the five normalization layers. Conventional batch normalization degrades the performance on the target domain, as we also observed in Table \ref{table__1plusplus}. Instance normalization introduces more noise during training and we observe that even the performance on source-only is lowest of the three normalization layers. Split batch normalization does not apply the same transformation to each domain and has the lowest gap reduction. Finally, we observe that the gap reduction using DAN ranks highest at $\gapredcustom$ points mIOU.

\subsection{Unsupervised adversarial domain adaptation} \label{results:UADA}
\begin{figure*}[tb]
	\centering
	\includegraphics[width=\linewidth]{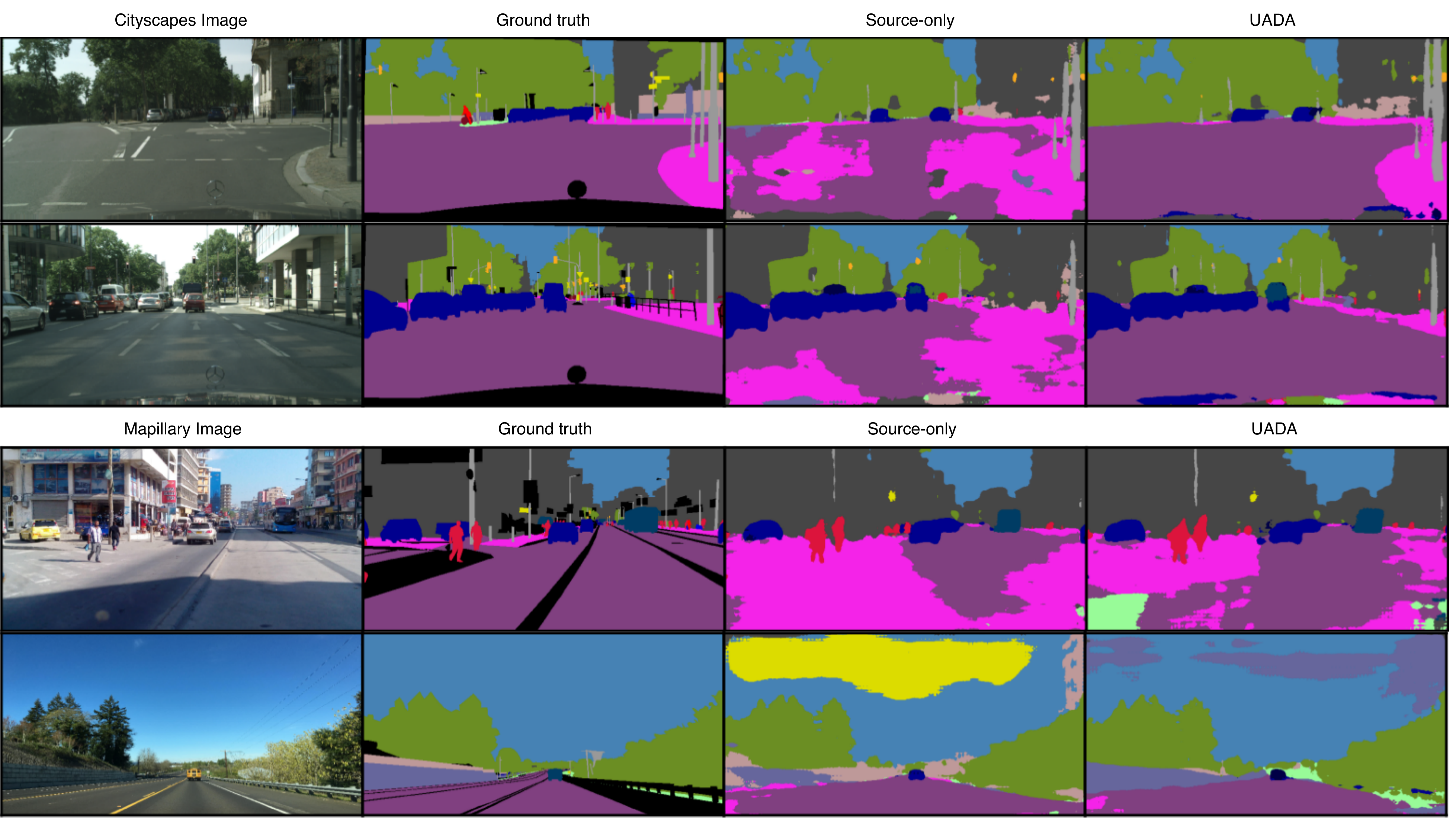}
	\caption{Example results for images from the Cityscapes and Mapillary validation sets. Per image, we show predictions from a model trained on source-only (GTA5) and UADA (\adapt{GTA5}{Cityscapes}) using DAN. The supplementary material contains more example results.}
	\label{comp1}
\end{figure*}

\begin{table}[t]
	\caption{\textbf{Hyperparameter analysis} Comparing target performance for ranging values of $\lambda$. CS abbreviates Cityscapes. Numbers represent target performance in units \% mIOU}
	\centering
	\setlength\tabcolsep{4pt}
	\begin{tabularx}{\linewidth}{Y|c|c|c|c|c}
		value of $\lambda$          & $10^{-5}$ & $10^{-4}$ & $10^{-3}$                  & $10^{-2}$                & $10^{-1}$  \Bs \\
		\hline 
		\hline 
		\adapt{GTA5}{CS}      &           & 33.6      & 34.6                       & \textbf{\uadacityscapes} & 34.6     \Ts \\
		\adapt{CS}{Mapillary} & 42.6      & 43.0      & \textbf{\uadatwomapillary} & 37.6                     &           \\
	\end{tabularx}
	\label{table__6lambda}
\end{table}

\begin{table}[t]
	\caption{\textbf{\adapt{GTA5}{Cityscapes}}. First row, source, indicates a model trained on GTA5 only. We also report performance on two unseen domains, Mapillary and Apolloscape. All units are \% mIOU.}
	\label{table__3uda1}
	\centering
	\setlength\tabcolsep{3pt}
	\begin{tabularx}{\linewidth}{Y||c|c|c|c}
		\tableheaderadapt{5}{GTA5}{Cityscapes}
		& \multicolumn{4}{c}{Tested on} \\ 
		& GTA5              & Cityscapes                    & Mapillary               & Apolloscape          \\
		Method      & (\textit{Source}) & (\textit{Target})             & (\textit{Unseen})       & (\textit{Unseen})    \Bs\\
		\cline{2-5}
		Source & $    62.2 $       & \gtacityscapes                & \gtamapillary           & \gtaapollo           \Ts\\
		UADA        & \textbf{\uadagta} & \textbf{    \uadacityscapes } & \textbf{\uadamapillary} & \textbf{\uadaapollo} \\	
	\end{tabularx} %
\end{table}

\begin{table}[t]
	\caption{\textbf{Comparison with three recent works in UADA.} The first column indicates at what level the adversary operates. Image size indicates number of pixels in height $\times$ width. Performance units are \% mIOU.}
	\label{table__5uda1_comparison}
	\centering
	\setlength\tabcolsep{2pt}
	\begin{tabularx}{\linewidth}{Y|c|c|c}
		\tableheaderadapt{3}{GTA5}{Cityscapes}
		Adversarial adaptation level              & Norm. & Image & Target            \\
		& layer & size & perf. \Bs                \\
		\hline 
		Representation \cite{FCNWild}             & No & \text{x} & 25.5 \Ts                 \\
		Representation and logit \cite{Zhang2017} & No & \text{x}   & 28.9                     \\
		Representation and output \cite{UnsupGAN} & No & $512\times1024$   & 37.1                     \\
		Representation [ours]                     & Yes& $384\times768$   & \textbf{\uadacityscapes} \\
	\end{tabularx}
\end{table}

\par 
In this experiment, we evaluate UADA using DAN. We consider two adaptation tasks, compare with related works and report results on unseen domains, Mapillary and Apolloscape. We found the values for $\lambda$ using a hyperparameter sweep and report results in Table \ref{table__6lambda}.

\par 
Table \ref{table__3uda1} reports the results for adapting from GTA5 to Cityscapes. These results show that we improve $\gapredcustom$ points mIOU on the target domain. A source-only model achieves $\gtagta$ \% mIOU on GTA5 and $\gtacityscapes$ \% mIOU on Cityscapes. UADA achieves $\uadagta$ \% mIOU on GTA5 and $\uadacityscapes$ \% mIOU on Cityscapes. Referring back to Figure \ref{retrieval_plot}, we observe that the retrieval curve for the adapted model (yellow) lies closer to the perfect alignment curve, which confirms that UADA aligns the representations. We plot segmentation outputs of both models on the Cityscapes validation set in Figure \ref{comp1}. We observe that the source-only model, trained on GTA5, makes incorrect predictions in large areas of the Cityscapes images. The plots for UADA clearly show less of these errors. In Table \ref{table__5uda1_comparison}, we compare our results to recent publications on this adaptation task. Despite using lower resolution, training UADA using DAN achieves the highest performance, $\uadacityscapes$ \% mIOU, on the target domain.

\par 
Table \ref{table__3uda1} shows that the performance improves on two unseen domains, Mapillary and Apolloscape. For Mapillary, the source-only model achieves $\gtamapillary$ \% mIOU and UADA improves the performance to $\uadamapillary$ \% mIOU. For Apolloscape, the source-only model achieves $\gtaapollo$ \% mIOU and UADA improves the performance to $\uadaapollo$ \% mIOU. These results show that by adapting from GTA5 to Cityscapes, the performance improves $\gapredcustommapillary$ points mIOU on Mapillary and $\gapredcustomapollo$ points mIOU on Apollo. Appendix \ref{comp1} shows segmentation outputs of both models on an unseen domain, Mapillary. Again on this unseen domain, we observe less incorrect predictions when comparing the unadapted with the adapted model.   

\par 
Finally, we reproduce the benefits of DAN on a second, real world adaptation task: from Cityscapes to Mapillary. Both domains are real and have less domain gap. Consequently, we expect a smaller improvement compared to synthetic to real adaptation. Table \ref{table__4uda2} reports the results for this adaptation task. We observe that the source domain improves from $\cityscapescityscapes$ to $\uadatwocityscapes$ \% mIOU. The target domain improves $\gapreduadatwo$ points from $\cityscapesmapillary$ to $\uadatwomapillary$ \% mIOU. For this second adaptation task, we evaluate the source-only model and the adapted model on the unseen domain, Apolloscape. The performance on Apolloscape improves $\gapreduadatwoapollo$ points from $\cityscapesapollo$ to $\uadatwoapollo$ \% mIOU. Although these improvements are less than adapting from a synthetic to a real domain, this experiment again shows the improved generalization capability for unseen domains, which is a chief aim in pattern recognition.
\begin{table}
	\caption{\textbf{\adapt{Cityscapes}{Mapillary}}. The first row, source, indicates a model trained on Cityscapes only. We also report performance on an unseen domain, Apolloscape. All units are \% mIOU.}
	\label{table__4uda2}
	\centering
	\begin{tabularx}{\linewidth}{c||Y|Y|Y}
		\tableheaderadapt{4}{Cityscapes}{Mapillary}
		& \multicolumn{3}{c}{Tested on}  \\ 
		            & Cityscapes                  & Mapillary                       & Apolloscape             \\
		Method      & (\textit{Source})           & (\textit{Target})               & (\textit{Unseen})      \Bs \\
		\cline{2-4}
		Source & \cityscapescityscapes       & \cityscapesmapillary            & \cityscapesapollo     \Ts  \\
		UADA        & \textbf{\uadatwocityscapes} & \textbf{    \uadatwomapillary } & \textbf{\uadatwoapollo} \\	
	\end{tabularx} %
\end{table}

\section{Discussion} \label{discussion}

\par 
Our experimental results demonstrate that UADA with our normalization layer improves segmentation performance on source, target and unseen domains. In this section, we highlight some points for discussion.
\par 
In training UADA, the $\lambda$ hyperparameter plays a critical role for the performance on the target domain. The $\lambda$ balances the segmentation and confusion loss, see Equation \ref{eq_alternating}. In Table \ref{table__6lambda}, we report the target performance on both adaptation tasks using different values for $\lambda$. We observe that 1) the performance drops more than $2$ points mIOU when we change $\lambda$ from the optimal value; 2) the optimal value differs between the adaptation tasks. However, in unsupervised domain adaptation, we would have no labels for the target domain to evaluate these numbers. One might thus argue that tuning the hyperparameters alludes to overfitting on the target labels. Therefore, we evaluate the model on unseen domains, i.e. the domains that we never used training nor setting the hyperparameters. For both adaptations tasks in Tables \ref{table__3uda1} and \ref{table__4uda2}, we observe an improvement in the performance on unseen domains. These improvements show that training with the optimal $\lambda$ value for the target domain also improves performance for unseen domains.

\par 
The domain classifier operates at the level of the representations, which might not suffice to reduce the domain adaptation gap. An adapted model from GTA5 to Cityscapes achieves $\uadacityscapes$ \% mIOU, while a model trained and evaluated on Cityscapes achieves $\cityscapescityscapes$ \% mIOU. This gap shows the need for further improvement. The first column in Table \ref{table__5uda1_comparison} shows at what level the adversaries operate. Contemporary work focuses on extending the adversaries to multiple levels, \cite{UnsupGAN,NoMore}. These works show a promising direction of research and we believe that our DAN will facilitate further improvements.


\section{Conclusion}
\par 
In this work, we present a Domain Agnostic Normalization (DAN) layer for unsupervised adversarial domain adaptation (UADA). Unlike conventional batch normalization, for which we show that it significantly reduces performance on UADA, our DAN layer learns normalization statistics only over the labeled domain and applies it to unlabeled and unseen domains. Consequently, DAN unlocks the benefits of normalization for UADA and we present in-depth experiments to demonstrate its advantage. Despite training at less than half of the original resolution, we surpass state-of-the-art performance on a common benchmark, i.e. adapting from GTA5 to Cityscapes, achieving $\uadacityscapes$ \% mIOU on the target domain. As a model trained with DAN remains domain agnostic when testing, we also report $\gapredcustommapillary$ and $\gapredcustomapollo$ points mIOU improvement on two unseen domains, Mapillary and Apolloscape, respectively. We reproduce these improvements on a second adaptation task, i.e. adapting from Cityscapes to Mapillary. Again, we show a performance improvement on an unseen domain, Apolloscape. All together, we demonstrate that using DAN allows for improved performance on unseen domains, which is the chief aim of pattern recognition.

{\small
\bibliographystyle{ieee}
\bibliography{library}
}

\end{document}